\documentclass[twocolumn,conference]{IEEEtran}
\usepackage[T1]{fontenc}
\usepackage[latin9]{inputenc}
\usepackage{amsmath}
\usepackage[unicode=true,
 bookmarks=true,bookmarksnumbered=true,bookmarksopen=true,bookmarksopenlevel=1,
 breaklinks=false,pdfborder={0 0 0},pdfborderstyle={},backref=false,colorlinks=false]
 {hyperref}
\hypersetup{pdftitle={Your Title},
 pdfauthor={Your Name},
 pdfborderstyle={},pdfpagelayout=OneColumn,pdfnewwindow=true,pdfstartview=XYZ,plainpages=false}
\usepackage{breakurl}

\makeatletter

\providecommand{\tabularnewline}{\\}

\makeatother

\begin{document}

\title{Hierarchical Convolutional-Deconvolutional Neural Networks for Automatic Liver and Tumor Segmentation}

\author{\IEEEauthorblockN{Yading~Yuan}\IEEEauthorblockA{Department of Radiation Oncology\\
 Icahn School of Medicine at Mount Sinai\\
 New York, NY, USA}}
\maketitle
\begin{abstract}
Automatic segmentation of liver and its tumors is an essential step
for extracting quantitative imaging biomarkers for accurate tumor
detection, diagnosis, prognosis and assessment of tumor response to
treatment. MICCAI 2017 Liver Tumor Segmentation Challenge (LiTS)\footnote{http://lits-challenge.com} provides
a common platform for comparing different automatic algorithms on
contrast-enhanced abdominal CT images in tasks including 1) liver
segmentation, 2) liver tumor segmentation, and 3) tumor burden estimation.
We participate this challenge by developing a hierarchical framework
based on deep fully convolutional-deconvolutional neural networks
(CDNN). A simple CDNN model is firstly trained to provide a quick
but coarse segmentation of the liver on the entire CT volume, then
another CDNN is applied to the liver region for fine liver segmentation.
At last, the segmented liver region, which is enhanced by histogram
equalization, is employed as an additional input to the third CDNN
for tumor segmentation. Jaccard distance is used as loss function
when training CDNN models to eliminate the need of sample re-weighting.
Our framework is trained using the 130 challenge training cases provided
by LiTS. The evaluation on the 70 challenge testing cases resulted
in a mean Dice Similarity Coefficient (DSC) of 0.963 for liver segmentation,
a mean DSC of 0.657 for tumor segmentation, and a root mean square
error (RMSE) of 0.017 for tumor burden estimation, which ranked our
method in the first, fifth and third place, respectively. 
\end{abstract}

\IEEEpeerreviewmaketitle{}

\section{Introduction}

Liver cancer is the second leading cause of global cancer mortality
(after lung cancer), and is one of the most rapidly increasing cancers
in terms of incidence and mortality worldwide and in the United States
\cite{Ferlay2015,Petrick2016}. Although contrast-enhanced computed
tomography (CT) has been widely used for liver cancer screening, diagnosis,
prognosis, and the assessment of its response to treatment, proper
interpretation of CT images is normally time-consuming and prone to
suffer from inter- and intra-observer variabilities. Therefore, computerized
analysis methods have been developed to assist radiologists and oncologists
for better interpretation of liver CT images.

Automatically segmenting liver and viable tumors from other tissue
is an essential step in quantitative image analysis of abdominal CT
images. However, automatic liver segmentation is a challenging task
due to the low contrast inside liver, fuzzy boundaries to its adjacent
organs and highly varying shape. Meanwhile, automatic tumor segmentation
on liver normally suffers from significant variety of appearance in
size, shape, location, intensity, textures, as well as the number
of occurrences. Although researchers have developed various methods
to conquer these challenges \cite{Heimann:2009,Yuan:2015,Linguraru:2012},
interactive approaches are still the only way to achieve acceptable
tumor segmentation.

In this paper, we present a fully automatic framework based on deep
fully convolutional-deconvolutional neural networks (CDNN) \cite{long2015fully,noh2015learning,yuan2017ieee}
for liver and liver tumor segmentation on contrast-enhanced abdominal
CT images. Similar to \cite{christ2016automatic}, our framework is
hierarchical and includes three steps. In the first step, a simple
CDNN model is trained to obtain a quick but coarse segmentation of
the liver on the entire 3D CT volume; then another CDNN is applied
to the liver region for fine liver segmentation; finally, the segmented
liver region is enhanced by histogram equalization and serves as an
additional input to the third CDNN for tumor segmentation. Instead
of developing sophisticated pre- and post-processing methods and hand-crafted
features, we focus on designing appropriate network architecture and
efficient learning strategies such that our framework can handle images
under various acquisition conditions.

\section{Dataset and Preprocessing}

Only LiTS challenge datasets were used for model training and validation.
The LiTS datasets consist of $200$ contrast-enhanced abdominal CT
scans provided by various clinical sites around the world, in which
$130$ cases were used for training and the rest $70$ for testing.
The datasets have significant variations in image quality, spatial
resolution and field-of-view, with in-plane resolution ranging from
$0.6\times0.6$ to $1.0\times1.0$ mm and slice thickness from $0.45$
to $6.0$ mm. Each axial slice has identical size of $512\times512$,
but the number of slices in each scan varies from $42$ to $1026$.

As for pre-processing, we simply truncated the voxel values of all
CT scans to the range of {[}-100, 400{]} HU to eliminate the irrelevant
image information. While a comprehensive 3D contextual information
could potentially improve the segmentation performance, due to the
limited hardware resource, it is infeasible to perform a fully 3D
CDNN on the volumetric CT scans in our experimental environment. Thus,
our CDNN model is based on 2D slice and the CT volume is processed
slice-by-slice, with the two most adjacent slices concatenated as
additional input channels to the CDNN model. Different resampling
strategies were applied at different hierarchical levels and will
be described below.

\section{Method}

\subsection{CDNN model}

Our CDNN model \cite{yuan2017ieee} belongs to the category of fully
convolutional network (FCN) that extends the convolution process across
the entire image and predicts the segmentation mask as a whole. This
model performs a pixel-wise classification and essentially serves
as a filter that projects the 2D CT slice to a map where each element
represents the probability that the corresponding input pixel belongs
to liver (or tumor). This model consists two pathways, in which contextual
information is aggregated via convolution and pooling in the convolutional
path and full image resolution is recovered via deconvolution and
up-sampling in the deconvolutional path. In this way, the CDNN model
can take both global information and fine details into account for
image segmentation.

We fix the stride as $1$ and use Rectified Linear Units (ReLUs) \cite{krizhevsky2012imagenet}
as the activation function for each convolutional/deconvolutional
layer. For output layer, we use sigmoid as the activation function.
Batch normalization is added to the output of every convolutional/deconvolutional
layer to reduce the internal covariate shift \cite{ioffe2015batch}.

We employ a loss function based on Jaccard distance proposed in \cite{yuan2017ieee}
in this study: 
\begin{equation}
L_{d_{J}}=1-\frac{\underset{i,j}{\sum}(t_{ij}p_{ij})}{\underset{i,j}{\sum}t_{ij}^{2}+\underset{i,j}{\sum}p_{ij}^{2}-\underset{i,j}{\sum}(t_{ij}p_{ij})},\label{eq:ja-loss}
\end{equation}
where $t_{ij}$ and $p_{ij}$ are target and the output of pixel $(i,\,j)$,
respectively. As compared to cross entropy used in the previous work
\cite{christ2016automatic,ronneberger2015u}, the proposed loss function
is directly related to image segmentation task because Jaccard index
is a commonly used metric to assess medical imaging segmentation.
Meanwhile, this loss function is well adapted to the problems with
high imbalance between foreground and background classes as it does
not require any class re-balancing. We trained the network using Adam
optimization \cite{kingma2014adam} to adjust the learning rate based
on the first and the second-order moments of the gradient at each
iteration. The initial learning rate was set as $0.003$.

In order to reduce overfitting, we added two dropout layers with $p=0.5$
- one at the end of convolutional path and the other right before
the last deconvolutional layer. We also employed two types of image
augmentations to further improve the robustness of the proposed model
under a wide variety of image acquisition conditions. One consists
of a series of geometric transformations, including randomly flipping,
shifting, rotating and scaling. The other type focuses on randomly
normalizing the contrast of each input channels in the training image
slices.Note that these augmentations only require little extra computation,
so the transformed images are generated from the original images for
every mini-batch within each iteration.

\subsection{Liver localization}

This step aims to locate the liver region by performing a fast but
coarse liver segmentation on the entire CT volume, thus we designed
a relatively simple CDNN model for this task. This model, named CDNN-I,
includes $19$ layers with $230,129$ trainable parameters and its
architectural details can be found in \cite{yuan2017ieee}. For each
CT volume, the axial slice size was firstly reduced to $128\times128$
by down-sampling and then the entire image volume was resampled with
slice thickness of $3$ mm. We found that not all the slices in a
CT volume were needed in training this CDNN model, so only the slices
with liver, as well as the $5$ slices superior and inferior to the
liver were included in the model training. For liver localization
and segmentation, the liver and tumor labels were merged as a single
liver label to provide the ground truth liver masks during model training.

During testing, the new CT images were pre-processed following the
same procedure as training data preparation, then the trained CDNN-I
was applied to each slice of the entire CT volume. Once all slices
were segmented, a threshold of $0.5$ was applied to the output of
CDNN and a 3D connect-component labeling was performed. The largest
connected component was selected as the initial liver region.

\subsection{Liver segmentation}

An accurate liver localization enables us to perform a fine liver
segmentation with more advanced CDNN model while reducing computational
time. Specifically, we firstly resampled the original image with slice
thickness of $2$ mm, then the bounding-box of liver was extracted
and expanded by $10$ voxels in $x,$ $y$ and $z$ directions to
create a liver volume of interest (VOI). The axial dimensions of VOI
were further adjusted to $256\times256$ either by down-sampling if
any dimension was greater than $256$, or by expanding in $x$ and/or
$y$ direction otherwise. All slices in the VOI were used for model
training.

The CDNN model used in the liver segmentation (named CDNN-II) includes
$29$ layers with about $5\,M$ trainable parameters. As compared
to CDNN-I, the size of $local\,receptive\,field$ (LRF), or filter
size, is reduced in CDNN-II such that the network can go deeper, i.e.
more number of layers, which allows applying more non-linearities
and being less prone to overfitting \cite{simonyan2014very}. Meanwhile,
the number of feature channels is doubled in each layer. Please refer
to \cite{yuan2017automatic} for more details.

During testing, liver VOI was extracted based on the initial liver
mask obtained in the liver localization step, then the trained CDNN-II
was applied to each slice in the VOI to yield a 3D probability map
of liver. We used the same post-processing as liver localization to
determine the final liver mask.

\subsection{Tumor segmentation}

The VOI extraction in tumor segmentation was similar to that in liver
segmentation, except that the original image resolution was used to
avoid potentially missing small lesions due to image blurring from
resampling. Instead of using all the slices in the VOI, we only collected
those slices with tumor as training data so as to focus the training
on the liver lesions and reduce training time. Besides the original
image intensity, a 3D regional histogram equalization was performed
to enhance the contrast between tumors and surrounding liver tissues,
in which only those voxels within the 3D liver mask were considered
in constructing intensity histogram. The enhanced image served as
an additional input channel to another CDNN-II model for tumor segmentation.
We found this additional input channel could further boost tumor segmentation
performance.

During testing, liver VOI was extracted based on the liver mask from
the liver segmentation step. A threshold of $0.5$ was applied to
the output of CDNN-II model and liver tumors were determined as all
tumor voxels within the liver mask.

\subsection{Implementation}

Our CDNN models were implemented with Python based on Theano \cite{team2016theano}
and Lasagne\footnote{http://github.com/Lasagne/Lasagne} packages.
The experiments were conducted using a single Nvidia GTX 1060 GPU
with 1280 cores and 6GB memory.

We used five-fold cross validation to evaluate the performance of
our models on the challenge training datasets. The total number of
epochs was set as $200$ for each fold. When applying the trained
models on the challenge testing datasets, a bagging-type ensemble
strategy was implemented to combine the outputs of six models to further
improve the segmentation performance \cite{yuan2017ieee}.

An epoch in training CDNN-I model for liver localization took about
$70$ seconds, but the average time per epoch became 610 seconds and
500 seconds when training CDNN-II models for liver segmentation and
tumor segmentation, respectively. This increase was primarily due
to larger slice size and more complicated CDNN models. Applying the
entire segmentation framework on a new test case was, however, very
efficient, taking about $33$ seconds on average ($8$, $8$ and $17$
s for liver localization, liver segmentation and tumor segmentation,
respectively).

\section{Results and Discussion}

We applied the trained models to the $70$ LiTS challenge test cases (team: deepX).
Based on the results from the challenge organizers, our method achieved
an average dice similarity coefficient (DSC) of $0.963$ for liver
segmentation, a DSC of $0.657$ for tumor segmentation, and a root
mean square error (RMSE) of $0.017$ for tumor burden estimation,
which ranked our method in the first, fifth and third place, respectively.
The complete evaluation results are shown in Table \ref{tab:Liver-segmentation-results}-\ref{tab:Tumor-burden-results}.

To summarize our work, we develop a fully automatic framework for
liver and its tumor segmentation on contrast-enhanced abdominal CT
scans based on three steps: liver localization by a simple CDNN model
(CDNN-I), liver fine segmentation by a deeper CDNN model with doubled
feature channels in each layer (CDNN-II), and tumor segmentation by
CDNN-II model with enhanced liver region as additional input feature.
Our CDNN models are fully trained in an end-to-end fashion with minimum
pre- and post-processing efforts.

While sharing some similarities with previous work such as U-Net \cite{ronneberger2015u}
and Cascaded-FCN \cite{christ2016automatic}, our CDNN model is different
from them in the following aspects: 1) The loss function used in CDNN
model is based on Jaccard distance that is directly related to image
segmentation task while eliminating the need of sample re-weighting;
2) Instead of recovering image details by long skip connections as
in U-Net, the CDNN model constructs a deconvolutional path where deconvolution
is employed to densify the coarse activation map obtained from up-sampling.
In this way, feature map concatenation and cropping are not needed.

Due to the limited hardware resource, training a complex CDNN model
is very time consuming and we had to restrict the total number of
epochs to $200$ in order to catch the deadline of LiTS challenge
submission. While upgrading hardware is clearly a way to speed up
the model training, we plan to improve our network architectures and
learning strategies in our future work such that the models can be
trained in a more effective and efficient way. Other post-processing
methods, such as level sets \cite{cha2016urinary} and conditional
random field (CRF) \cite{chen2014semantic}, can also be potentially
integrated into our model to further improve the segmentation performance.

\begin{table}[htbp]
\caption{Liver segmentation results (deepX) on LiTS testing cases\label{tab:Liver-segmentation-results}}

\centering{}%
\begin{tabular}{|c|c|c|c|c|c|c|}
\hline 
Dice / case  & Dice global  & VOE  & RVD  & ASSD  & MSSD  & RMSD\tabularnewline
\hline 
\hline 
$0.9630$  & $0.9670$  & $0.071$  & $-0.010$  & $1.104$  & $23.847$  & $2.303$\tabularnewline
\hline 
\end{tabular}
\end{table}

\begin{table}[htbp]
\caption{Tumor segmentation results (deepX) on LiTS testing cases\label{tab:Tumor-segmentation-results}}

\centering{}%
\begin{tabular}{|c|c|c|c|c|c|c|}
\hline 
Dice / case  & Dice global  & VOE  & RVD  & ASSD  & MSD  & RMSD\tabularnewline
\hline 
\hline 
$0.6570$  & $0.8200$  & $0.378$  & $0.288$  & $1.151$  & $6.269$  & $1.678$\tabularnewline
\hline 
\end{tabular}
\end{table}

\begin{table}[htbp]
\caption{Tumor burden results (deepX) on LiTS testing cases\label{tab:Tumor-burden-results}}

\centering{}%
\begin{tabular}{|c|c|}
\hline 
RMSE  & Max Error\tabularnewline
\hline 
\hline 
$0.0170$  & $0.0490$\tabularnewline
\hline 
\end{tabular}
\end{table}


\begin{thebibliography}{10}
\bibitem{Ferlay2015} J.~Ferlay, I.~Soerjomataram, R.~Dikshit,
S.~Eser, C.~Mathers, M.~Robelo, D.~Parkin, D.~Forman, and F.~Bray,
``Cancer incidence and mortality worldwide: sources, methods and
major patterns in {GLOBCAN 2012},'' \emph{Int. J. Cancer}, vol.
136, no.~5, pp. E359\textendash 86, 2015.

\bibitem{Petrick2016} J.~Petrick, M.~Braunlin, M.~Laversanne,
P.~Valery, F.~Bray, and K.~McGlynn, ``International trends in
liver cancer incidence, overall and by histologic subtype, 1978-2007,''
\emph{Int. J. Cancer}, vol. 137, no.~7, pp. 1534\textendash 45, 2016.

\bibitem{Heimann:2009} T.~Heimann, B.~van Ginneken, M.~A. Styner,
Y.~Arzhaeva, V.~Aurich, C.~Bauer, A.~Beck, C.~Becker, R.~Beichel,
G.~Bekes, F.~Bello, G.~Binnig, H.~Bischof, A.~Bornik, P.~M.~M.
Cashman, Y.~Chi, A.~Cordova, B.~M. Dawant, M.~Fidrich, J.~D.
Furst, D.~Furukawa, L.~Grenacher, J.~Hornegger, D.~Kainmuller,
R.~I. Kitney, H.~Kobatake, H.~Lamecker, T.~Lange, J.~Lee, B.~Lennon,
R.~Li, S.~Li, H.-P. Meinzer, G.~Nemeth, A.-M.~R. D.~S.~Raicu,
E.~M. van Rikxoort, M.~Rousson, L.~Rusko, K.~A. Saddi, G.~Schmidt,
D.~Seghers, A.~Shimizu, P.~Slagmolen, E.~Sorantin, G.~Soza, R.~Susomboon,
J.~M. Waite, A.~Wimmer, and I.~Wolf, ``Comparison and evaluation
of methods for liver segmentation from {CT} datasets,'' \emph{IEEE
Trans. Med. Imaging}, vol.~28, pp. 1251\textendash 1265, 2009.

\bibitem{Yuan:2015} Y.~Yuan, M.~Chao, R.~Sheu, K.~Rosenzweig,
and Y.~Lo, ``Tracking fuzzy borders using geodesic curves with application
to liver segmentation on planning {CT},'' \emph{Med. Phys.}, vol.~42,
pp. 4015\textendash 26, 2015.

\bibitem{Linguraru:2012} M.~G. Linguraru, W.~J. Richbourg, J.~Liu,
J.~M. Watt, V.~Pamulapati, S.~Wang, and R.~M. Summers, ``Tumor
burden analysis on computed tomography by automated liver and tumor
segmentation,'' \emph{IEEE Trans. Med. Imaging}, vol.~31, pp. 1965\textendash 1976,
2012.

\bibitem{long2015fully} J.~Long, E.~Shelhamer, and T.~Darrell,
``Fully convolutional networks for semantic segmentation,'' in \emph{Proc.
CVPR}, 2015, pp. 3431\textendash 3440.

\bibitem{noh2015learning} H.~Noh, S.~Hong, and B.~Han, ``Learning
deconvolution network for semantic segmentation,'' in \emph{Proc.
ICCV 2015}, 2015, pp. 1520\textendash 1528.

\bibitem{yuan2017ieee} Y.~Yuan, M.~Chao, and Y.-C. Lo, ``Automatic
skin lesion segmentation using deep fully convolutional networks with
{Jaccard distance},'' \emph{IEEE Trans. Med. Imaging}, vol.~36, pp. 1876\textendash 1886, 2017.

\bibitem{christ2016automatic} {Christ, Patrick Ferdinand and Elshaer,
Mohamed Ezzeldin A and Ettlinger, Florian and Tatavarty, Sunil and
Bickel, Marc and Bilic, Patrick and Rempfler, Markus and Armbruster,
Marco and Hofmann, Felix and DAnastasi, Melvin and others}, ``{Automatic
liver and lesion segmentation in CT using cascaded fully convolutional
neural networks and 3D conditional random fields},'' in \emph{International
Conference on Medical Image Computing and Computer-Assisted Intervention}.\hskip
1em plus 0.5em minus 0.4em\relax Springer, 2016, pp. 415\textendash 423.

\bibitem{krizhevsky2012imagenet} A.~Krizhevsky, I.~Sutskever, and
G.~E. Hinton, ``Imagenet classification with deep convolutional
neural networks,'' in \emph{Adv. Neural Inf. Process. Sys.}, 2012,
pp. 1097\textendash 1105.

\bibitem{ioffe2015batch} S.~Ioffe and C.~Szegedy, ``Batch normalization:
Accelerating deep network training by reducing internal covariate
shift,'' \emph{arXiv preprint arXiv:1502.03167}, 2015.

\bibitem{ronneberger2015u} O.~Ronneberger, P.~Fischer, and T.~Brox,
``U-net: Convolutional networks for biomedical image segmentation,''
in \emph{Proc. MICCAI 2015}.\hskip 1em plus 0.5em minus 0.4em\relax
Springer, 2015, pp. 234\textendash 241.

\bibitem{kingma2014adam} D.~Kingma and J.~Ba, ``Adam: A method
for stochastic optimization,'' \emph{arXiv preprint arXiv:1412.6980},
2014.

\bibitem{simonyan2014very} K.~Simonyan and A.~Zisserman, ``Very
deep convolutional networks for large-scale image recognition,''
\emph{arXiv preprint arXiv:1409.1556}, 2014.

\bibitem{yuan2017automatic} Y.~Yuan, and Y.-C. Lo, ``Improving
dermoscopic image segmentation with enhanced convolutional-deconvolutional
networks,'' \emph{arXiv preprint arXiv:1709.09780}, 2017.

\bibitem{team2016theano} {The Theano Development Team}, R.~Al-Rfou,
G.~Alain, A.~Almahairi, C.~Angermueller, D.~Bahdanau, N.~Ballas,
F.~Bastien, J.~Bayer, A.~Belikov \emph{et~al.}, ``Theano: A python
framework for fast computation of mathematical expressions,'' \emph{arXiv
preprint arXiv:1605.02688}, 2016.

\bibitem{cha2016urinary} K.~H. Cha, L.~Hadjiiski, R.~K. Samala,
H.-P. Chan, E.~M. Caoili, and R.~H. Cohan, ``{Urinary bladder
segmentation in CT urography using deep-learning convolutional neural
network and level sets},'' \emph{Med. Phys.}, vol.~43, no.~4,
pp. 1882\textendash 1896, 2016.

\bibitem{chen2014semantic} L.-C. Chen, G.~Papandreou, I.~Kokkinos,
K.~Murphy, and A.~L. Yuille, ``{Semantic image segmentation with
deep convolutional nets and fully connected CRFs},'' \emph{arXiv
preprint arXiv:1412.7062}, 2014.
\end{thebibliography}
\end{document}